\documentclass[11pt,a4paper]{article}
\usepackage[hyperref]{acl2021}
\usepackage{times}
\usepackage{latexsym}

\usepackage{microtype}

\aclfinalcopy 
\def\aclpaperid{1905} 

\usepackage{amsmath,amsfonts,amssymb,bm,bbm,mathtools,upgreek}
\usepackage{graphicx}
\usepackage{xcolor}
\usepackage{caption}
\usepackage{subcaption}
\usepackage[all]{nowidow}
\usepackage{booktabs}
\usepackage{cleveref}
\crefformat{section}{\S#2#1#3}
\AtBeginDocument{%
  \crefformat{subsection}{\S#2#1#3}
  \crefformat{subsubsection}{\S#2#1#3}
}
\usepackage[multiple]{footmisc}

\def\attention{\mathrm{attn}}
\def\softmax{\mathrm{softmax}}
\def\min{\mathrm{min}}
\def\MLP{\mathrm{MLP}}
\def\MHA{\mathrm{MHA}}
\def\BERT{\mathrm{BERT}}
\def\CLS{\mathrm{CLS}}
\def\BERTCLS{\BERT_\CLS}
\def\head{\mathrm{head}}
\def\log{\mathrm{log}}
\def\SS{\textsc{S}}
\def\RR{\textsc{R}}
\def\NN{\textsc{N}}
\def\SSS{\textsc{Supported}}
\def\RRR{\textsc{Refuted}}
\def\NNN{\textsc{NotEnoughInfo}}
\def\R{\mathbb{R}}
\def\numDoc{K}
\def\numEvi{M}
\def\negRatio{r}
\def\numHead{n}
\def\dimHid{d_h}
\def\dimHead{\frac{\dimHid}{\numHead}}
\def\N{\mathcal{N}}
\def\Z{\mathcal{Z}}
\def\Y{\mathcal{Y}}
\def\D{\mathcal{D}}
\def\a{\mathbf{a}}
\def\e{\mathbf{e}}
\def\s{\mathbf{s}}
\def\c{\mathbf{c}}
\def\h{\mathbf{h}}
\def\g{\mathbf{g}}
\def\W{\mathbf{W}}
\def\Q{\mathbf{Q}}
\def\K{\mathbf{K}}
\def\V{\mathbf{V}}
\def\E{\mathbf{E}}
\def\F{\mathbf{F}}
\def\G{\mathbf{G}}
\def\H{\mathbf{H}}
\def\pS{p_{\paramS}}
\def\pzcej{\pS(\hat{z}|c,e_j)}
\def\sumZinZ{\sum_{z\in\Z}}
\def\LossSelectionj{\mathcal{L}_{e_j}}
\def\paramS{\bm{\phi}}
\def\EqLossSelectionj{\LossSelectionj(\paramS)=-\sumZinZ\indicator\{\hat{z}=z\}\log\pzcej}
\def\EqsSentSelection{
  \pzcej &= \softmax(\MLP(\e_j)),\label{eq:softmax-ej}\numberthis\\
  \e_j   &= \BERTCLS(c,e_j).
}
\def\eviset{\{e_j\}_{j=1}^M}
\def\evivecset{\{\e_j\}_{j=1}^M}
\def\pP{p_{\paramP}}
\def\pyce{\pP(\hat{y}|c,\eviset)}
\def\sumYinY{\sum_{y\in\Y}}
\def\indicator{\mathbbm{1}}
\def\paramP{\bm{\theta}}
\def\LossPrediction{\mathcal{L}_p}
\def\EqLossPrediction{\LossPrediction(\paramP)=-\sumYinY\beta_y\indicator\{\hat{y}=y\}\log\pyce}
\def\EqAvec{\a = \MHA(\Q = \c, \K = \E, \V = \E)}
\def\EqsMHA{
  & \hspace{-.3em} \MHA(\Q, \K, \V) = [\head_1, \ldots, \head_n]\W^O, \numberthis\\
  & \hspace{-.3em} \head_i          = \attention(\Q\W_i^Q, \K\W_i^K, \V\W_i^V),\label{eq:head} \numberthis
}
\def\EqsValueOnly{
 \head_i &= \attention(\Q\W_i^Q, \K\W_i^K, \widetilde{\V}\W_i^V),\label{eq:headv} \numberthis\\
 \widetilde{\V} &= \V + \s\odot\V \label{eq:gate}\numberthis,
}

\def\EqLossJoint{\min_{\paramP,\paramS}\;\mathcal{L} = \LossPrediction + \lambda\sum_{j=1}^M\LossSelectionj}
\def\EqsClaimEvidVecs{
  \c   &= \BERTCLS(c), \\
  \e_j &= \BERTCLS(c,e_j).
}
\def\QKgamma{\frac{\Q\K^\top}{\raisebox{0.5ex}{$\gamma$}}}
\def\scorevec{\s = [s_1, \ldots, s_M]}

\def\hbeta{\hat{\beta}}
\def\EqNormCW{
  \beta_y = \frac{\hbeta_y}{\sumYinY\hbeta_y}, \quad \hbeta_y = \frac{N}{|\Y|\times N_y},\nonumber
}
\def\EqHj{\H_j = [\h_{j,1}, ..., \h_{j,L}]}
\def\Hset{\{\H_j\}_{j=1}^M}
\def\Gset{\{\G_j\}_{j=1}^M}
\def\gCLSset{\{\g_{j,1}\}_{j=1}^M}
\def\EqResConTok{\G = \H + \MHA(\H)}
\def\EqResConSent{\E = \F + \MHA(\F)}
\def\sdag{^\dagger}
\def\sddag{^\ddagger}
\def\sclub{^\clubsuit}
\def\sdia{^\diamondsuit}
\def\sspade{^\spadesuit}
\def\LtimesM{\thickmuskip=0mu\medmuskip=0mu\thinmuskip=0mu L \times M}
\newcommand\numberthis{\addtocounter{equation}{1}\tag{\theequation}}
\newcommand\T{\rule{0pt}{2.5ex}}
\newcommand\B{\rule[-1.2ex]{0pt}{0pt}}
\definecolor{goldenbrown}{rgb}{0.6, 0.4, 0.08}
\def\SSSC{\color{blue}{\textbf{\SSS}}}
\def\RRRC{\color{red}{\textbf{\RRR}}}
\def\NNNC{\color{goldenbrown}{\textbf{\NNN}}}

\title{A Multi-Level Attention Model for Evidence-Based Fact Checking}

\author{Canasai Kruengkrai \hspace{1.6em}
  Junichi Yamagishi \hspace{1.6em}
  Xin Wang \\
  National Institute of Informatics, Japan \\
 \texttt{\{canasai,jyamagishi,wangxin\}@nii.ac.jp} \\
}

\date{}

\begin{document}


\maketitle

\begin{abstract}
  Evidence-based fact checking aims to verify the truthfulness of a claim against evidence extracted from textual sources.
  Learning a representation that effectively captures relations between a claim and evidence can be challenging.
  Recent state-of-the-art approaches have developed increasingly sophisticated models based on graph structures.
  We present a simple model that can be trained on sequence structures.
  Our model enables inter-sentence attentions at different levels and can benefit from joint training.
  Results on a large-scale dataset for Fact Extraction and VERification (FEVER) show that our model outperforms the graph-based approaches
  and yields 1.09\% and 1.42\% improvements in label accuracy and FEVER score, respectively, over the best published model.\footnote{
    The code and model checkpoints are available at: \url{https://github.com/nii-yamagishilab/mla}.
  }
\end{abstract}

\section{Introduction}

False or misleading claims spread through online media faster and wider than the truth~\cite{Vosoughi18}.
False claims can occur in many different forms, e.g., fake news, rumors, hoaxes, propaganda, etc.
Identifying false claims that are likely to cause harm in the real world is important.
Generally, claims can be categorized into two types: verifiable and unverifiable.
Verifiable claims can be confirmed to be true or false as guided by evidence from credible sources,
while unverifiable claims cannot be confirmed due to insufficient information.

\begin{figure}[!t]
  \centering
  \includegraphics[width=0.95\columnwidth]{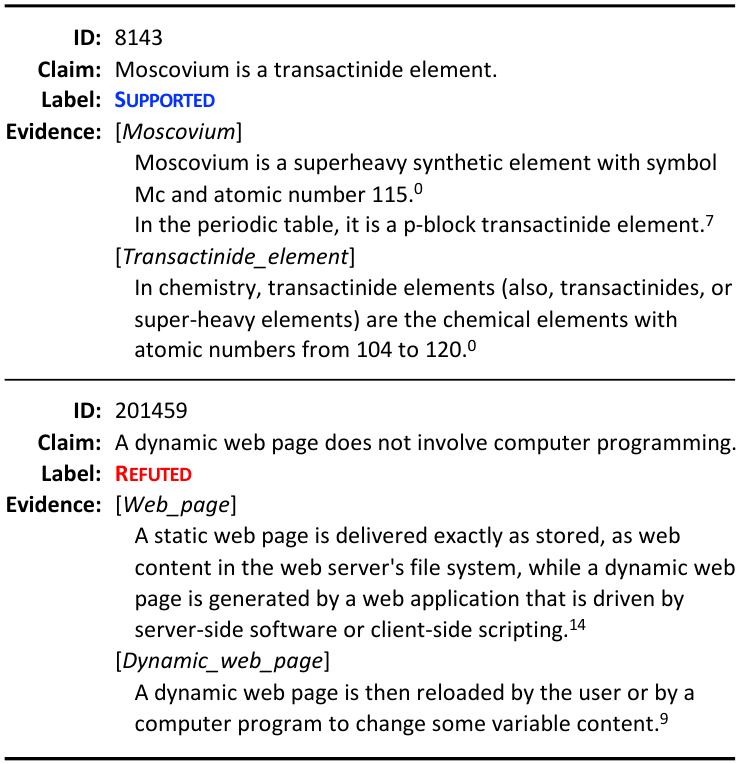}
  \caption{
    Examples from the FEVER dev set,
    where true evidence sentences are present in the selected sentences,
    and veracity relation labels are correctly predicted by our proposed model.
    Wikipedia article titles are in [\textit{italics}].
    Superscripts indicate the positions of the sentences in each article.
  }\label{fig:ex}
\end{figure}

Verifying the truthfulness of a claim with respect to evidence can be
regarded as a special case of recognizing textual entailment
(RTE)~\cite{Dagan06} or natural language inference
(NLI)~\cite{Bowman15}, where the premise (evidence) is not given.
Thus, the task of claim verification is
to first retrieve documents relevant to a given claim from textual sources,
then select sentences likely to contain evidence,
and finally assign a veracity relation label to support or refute the claim.
For example, the false claim
``\textit{Rabies is a foodborne illness.}''
can be refuted by the evidence
``\textit{Rabies is spread when an infected animal scratches or bites another animal or human.}''
extracted from the Wikipedia article ``\textit{Rabies}''.
Figure~\ref{fig:ex} shows other examples that require multiple evidence sentences to support or refute claims.
All of these claims are taken from a benchmark dataset for Fact Extraction and VERification (FEVER)~\cite{FEVER1}.
A key challenge is to obtain a representation for claim and evidence sentences that can effectively capture relations among them.

Recent state-of-the-art approaches have attempted to meet this challenge by applying graph-based neural networks~\cite{Kipf17,Velickovic18}.
For example, \newcite{GEAR19} regard an evidence sentence as a graph node,
while \newcite{KGAT20} use a more fine-grained node representation based on token-level attention.
\newcite{DREAM20} use semantic role labeling (SRL) to build a graph structure,
where a node can be a word or a phrase depending on the SRL's outputs.

In this paper, we argue that such sophisticated graph-based approaches may be unnecessary for the claim verification task.
We propose a simple model that can be trained on a sequence structure.
We also observe mismatches between training and testing.
At test time, the model predicts the veracity of a claim based on
retrieved documents and selected sentences, which contain prediction errors,
while at training time, only ground-truth documents and true evidence sentences are available.
We empirically show that our model, trained with a method that helps reduce training-test discrepancies, outperforms the graph-based approaches.

In addition, we observe that most of the previous work neglects sentence-selection labels when training veracity prediction models.
Thus, we propose leveraging those labels to further improve veracity relation prediction through joint training.
Unlike previous work that jointly trains two models~\cite{YinRoth18,LiSciFact20,Hidey20,Nie20},
our approach is still a pipeline process
where only a subset of  potential candidate sentences produced by \textit{any} sentence selector can be used for joint training.
This approach makes it possible to explore different sentence-selection models trained with different methods.

Our contributions are as follows.
We develop a method for mitigating training-test discrepancies by using a mixture of predicted and true examples for training.
We propose a multi-level attention (MLA) model that enables token- and sentence-level self-attentions and that benefits from joint training.
Experiments on the FEVER dataset show that MLA outperforms all the published models, despite its simplicity.

\section{Background and related work}

\subsection{Problem formulation}

The input of our task is a claim and a collection of Wikipedia articles $\D$.
The goal is to extract a set of evidence sentences from $\D$ and
assign a veracity relation label $\medmuskip=0mu y \in \Y=\{\SS,\RR,\NN\}$
to a claim with respect to the evidence set,
where
$\SS$ = $\SSS$, $\RR$ = $\RRR$, and $\NN$ = $\NNN$.
The definition of our labels is identical to that of the FEVER Challenge~\cite{FEVER1}.

\subsection{Overview of evidence-based fact checking}

\begin{figure}[!t]
  \centering
  \includegraphics[width=0.76\columnwidth]{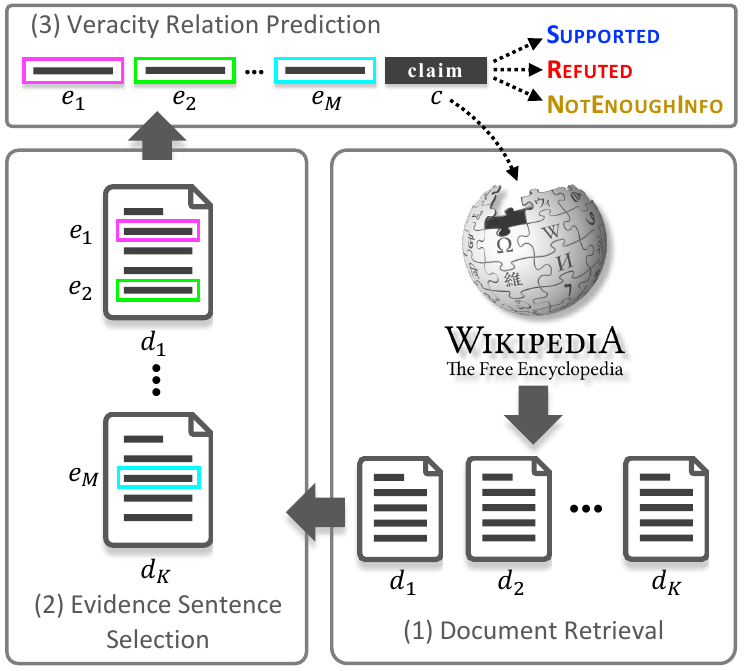}
  \caption{Process of evidence-based fact checking:
    retrieving documents relevant to a given claim from Wikipedia,
    selecting sentences likely to contain evidence,
   and predicting a veracity relation label based on selected sentences.
  }
  \label{fig:pipeline}
\end{figure}

The process of evidence-based fact checking, shown in Figure~\ref{fig:pipeline}, commonly involves the following three subtasks.

\subsubsection*{Document retrieval}

\noindent
Given a claim, the task is to retrieve the top $\numDoc$ relevant documents from $\D$.
\newcite{FEVER1} suggest using the document retriever from
DrQA~\cite{DrQA17}, which ranks documents on the basis of the term
frequency-inverse document frequency (TF-IDF) model with
unigram-bigram hashing.
\newcite{Athene18} use a hybrid approach that combines search results from the \mbox{MediaWiki} API and the results of using exact matching on all Wikipedia article titles.
In this paper, our main focus is to improve evidence sentence selection and veracity relation prediction,
so we directly use the document retrieval results from~\newcite{Athene18}.
This allows us to fairly compare our model with a series of previous methods~\cite{Soleimani19,GEAR19,KGAT20,Coref20}
that also rely on \newcite{Athene18}'s results.

\subsubsection*{Evidence sentence selection}

\noindent
The task is to select the top $\numEvi$ sentences from the retrieved documents.
\newcite{FEVER1} again use the TF-IDF model to rank sentences similar to a given claim.
\newcite{Nie19,Athene18} use the enhanced sequential inference model (ESIM)~\cite{ESIM17} to encode and align a claim-sentence pair.
\newcite{KGAT20,Athene18} use a pairwise hinge loss to rank sentences,
while~\newcite{Soleimani19} explore both pointwise and pairwise losses and suggest selecting difficult negative examples for training.
The pairwise hinge loss aims to maximize the margin between the scores of positive and negative examples,
while the pointwise loss is a vanilla cross-entropy loss.
Our model uses a pointwise loss trained with examples sampled from both ground-truth and predicted documents.

\subsubsection*{Veracity relation prediction}

\noindent
Given a claim and a set of $\numEvi$ selected sentences, the task is to predict their veracity relation label $y$.
Previous work on the FEVER Challenge modified existing RTE/NLI models to
deal with multiple sentences~\cite{Nie19,Yoneda18,Athene18,FEVER1},
used heuristic rules to combine predictions from individual
claim-sentence pairs~\cite{Malon18}, or concatenated all sentences~\cite{Stammbach19}.
A line of recent work has applied graph-based neural networks~\cite{GEAR19,KGAT20,DREAM20}.
Our model is simply trained on linear sequences by using self- and cross-attention to learn inter-sentence interactions.

\subsection{Pre-trained language models}

A key to the success of state-of-the-art approaches is the use of pre-trained language models~\cite{BERT19,RoBERTa19,XLNet19,ALBERT20}.
Here, we use BERT~\cite{BERT19}, a bidirectional transformer~~\cite{Vaswani17}, to obtain the vector representation of a token sequence.
Each BERT layer transforms an input token sequence (one or two sentences) by using self-attention.
The first hidden state vector of the final layer represents a special classification token (CLS), which can be used in downstream tasks.
We denote the above process by $\BERTCLS(\cdot) \in \R^{\dimHid}$,
where $\dimHid$ means the dimensionality of BERT hidden state vectors.

\section{Proposed method}

In this section, we describe our contributions, including
(1) our method for training the sentence-selection model and
(2) our veracity prediction model that can be extended with inter-sentence attentions and joint training.

\subsection{Learning to select sentences from  mixed ground-truth and retrieved documents}\label{sec:sent-selection}

Our goal is to select a subset of evidence sentences from all candidate sentences in the retrieved documents.
We consider this task to be a binary classification problem that takes
as input a pair of a claim $c$ and a candidate sentence $e_j$ and maps
it to the output $z \in \Z = \{-1, +1\}$, where $+1$ indicates an
evidence sentence and $-1$ otherwise.
We train our sentence-selection model by minimizing the standard
cross-entropy loss for each example:
\begin{equation}\label{eq:loss-selection}
  \EqLossSelectionj,
\end{equation}
where $\indicator\{\cdot\}$ is the indicator function, and $\pS$ is the
probability distribution of the two classes generated by our model.
We compute $\pS$ by applying a multi-layer perceptron (MLP) to the vector representation of $e_j$ followed by a softmax function: 
{\setlength{\jot}{0.5\baselineskip}
\begin{align*}
  \EqsSentSelection
\end{align*}
}%
The MLP contains two affine transformations that map $\e_j$ to the output space.
Feeding the pair of $c$ and $e_j$ to BERT allows us to obtain hidden state vectors that capture interactions between $c$ and $e_j$ at the token level.
This is due to the self-attention mechanism inside the BERT layers.
We expect the final hidden state vector of the CLS token (i.e., $\e_j$) to encode useful information from $e_j$ with respect to $c$.
The parameters $\paramS$ include those in MLP and BERT.

Training our model seems straightforward.
However, two technical issues exist.
First, each document typically contains one or two (or no) evidence sentences.
Training with a few positive examples (i.e., evidence sentences)
against all negative examples (i.e., non-evidence sentences) may be
neither efficient nor effective.
\newcite{Soleimani19} use hard negative mining (HNM) to repeatedly select a subset of difficult negative examples for training their sentence selector.
Second, at test time, the model must examine all candidate sentences in the relevant documents returned by the document retriever.
However, at training time, the model has no chance to learn the
characteristics of non-evidence sentences in the irrelevant but
highly ranked documents if only the ground-truth documents are used.

We propose to mitigate the aforementioned issues by using both the ground-truth and retrieved documents to create negative examples for a claim.
First, we randomly choose $\negRatio$ non-evidence sentences from each ground-truth document, where $r$ is twice the number of true evidence sentences.
Then, we sample two other non-evidence sentences from each retrieved document.
For positive examples, we use the true evidence sentences in the ground-truth documents.
Our scheme is more efficient than HNM of~\newcite{Soleimani19}.
At test time, we select the top $\numEvi$ sentences according to the probabilities assigned to the positive class.

\subsection{Multi-level attention and joint training for veracity relation prediction}

Training-test discrepancies also occur in veracity relation prediction.
At test time, the model predicts the veracity of a claim on the basis of the predicted evidence sentences.
At training time, only true evidence sentences are available for $\SSS$ and $\RRR$, but not for $\NNN$. 
In other words, we have no example sentences that more or less relate
to a claim but may not be sufficient to support or refute the claim
to train the model.
\newcite{FEVER1} simulate training examples for $\NNN$
by sampling a sentence from the highest-ranked page returned by the
document retriever.

We propose to reduce this discrepancy by using a mixture of true and predicted evidence sentences for training.
First, we pair each claim with a list of the top $\numEvi$ predicted sentences obtained through a sentence selector.
At training time, we then prepend the true evidence sentences (if available) to the list and keep the number of all the sentences at most $\numEvi$.\footnote{
  True evidence sentences may already exist in the list because the sentence selector can correctly identify them.
  }
At test time, we use the top $\numEvi$ predicted sentences without requiring a predefined threshold to filter them.
This is in contrast to previous work~\cite{GEAR19,SRMRS19,Wadden20} and helps reduce engineering effort.
Our example sentences for $\NNN$ are from the sentence selector, not from the document retriever as in~\cite{FEVER1}.
We expect our training examples to be similar to what our model may encounter at test time.

On the basis of the above scheme, each example is a pair of a claim $c$ and a set of evidence sentences $\eviset$.
Our goal is to predict the veracity relation label $\medmuskip=0mu y \in \Y=\{\SS, \RR, \NN\}$.
We train our veracity prediction model by minimizing the class-weighted cross-entropy loss for each example:
\begin{equation}\label{eq:loss-prediction}
  \EqLossPrediction,
\end{equation}%
where $\beta_y$ is the class weight for dealing with the class imbalance problem (detailed in Section~\ref{sec:training-details}).
Similar to Eq.~(\ref{eq:softmax-ej}), we compute the probability distribution $\pP$ of veracity relation labels as:
\begin{equation}\label{eq:pyce}
  \pyce = \softmax(\MLP(\a)).
\end{equation}
Here, $\a$ is the vector representation of aggregated evidence
about a claim that is obtained through the multi-head attention (MHA) function:
\begin{equation}\label{eq:a}
\thickmuskip=4mu \EqAvec,
\end{equation}
where $\c$ is the claim vector, $\E$ is the set of evidence vectors $\evivecset$,
and $\Q$, $\K$, $\V$ denote the query, keys, and values, respectively.
All the claim and evidence vectors are derived from BERT:
{\setlength{\jot}{0.5\baselineskip}
\begin{align*}
  \EqsClaimEvidVecs
\end{align*}
}%
The parameters $\paramP$ are those in MLP, MHA, and BERT.

Now let us explain the MHA function, because we use and/or modify it in other components.
The MHA function is based on the scaled dot-production attention~\cite{Vaswani17}:
\begin{equation}\label{eq:attn}
  \attention(\Q, \K, \V) = \softmax\Big(\QKgamma\Big)\V, \\
\end{equation}
where $\gamma=\sqrt{\dimHid/\numHead}$ is the scaling factor.
The above function is the weighted sum of the values (i.e., the evidence vectors), where the
weight assigned to each value is the result of applying a softmax function
to the scaled dot products between the query (i.e., the claim vector) and the keys (i.e., the evidence vectors).

The MHA function contains a number of parallel heads (i.e., attention layers).
We expect each head to capture different aspects of the input.
We achieve this by linearly projecting $\Q$, $\K$, and $\V$ to new representations and feeding them to the scaled dot-product attention.
Specifically, the MHA function is given by:
{\setlength{\jot}{0.5\baselineskip}
\begin{align*}
  \EqsMHA
\end{align*}
}%
where $\numHead$ is the number of parallel heads,
and $\W_i^Q$, $\W_i^K$, $\thickmuskip=3mu\W_i^V \in \R^{\dimHid\times\dimHead}$; $\thickmuskip=3mu\W^O \in \R^{\dimHid\times\dimHid}$ are the weight matrices of the linear projections.

\subsubsection*{Inter-sentence attentions}

\noindent
Although Eq.~(\ref{eq:a}) helps aggregate the evidence from multiple selected
sentences, our model still has no mechanism to learn interactions among these sentences.
Unlike previous work that uses graph-based attention~\cite{GEAR19,KGAT20,DREAM20},
our main tool is just the described MHA function.

Let $\EqHj$ be a sequence of the hidden state vectors of $e_j$ generated by BERT, where $L$ is the maximum sequence length.
Let $\H$ be the concatenation of all the sequences $\Hset$.
We obtain a new representation $\G$ of the concatenated sequence by
applying a residual connection between $\H$ and token-level self-attention:
\begin{equation}\label{eq:attn-tok}
\EqResConTok,
\end{equation}
where $\MHA(\cdot)$ is a simplified MHA function with one argument because $\Q$, $\K$, and $\V$ all come from the same $\H$.

In practice, we also add the static (sinusoid) positional encodings
(PE) to the input of MHA.\footnote{
During development, we tried the other basic components, i.e., layer normalization and position-wise feed-forward, but found it yielded no improvements in our task.}
We adopt this procedure from the original Transformer's sub-layer~\cite{Vaswani17}.
The computation cost of Eq.~(\ref{eq:attn-tok}) is not high.
Concretely, let $L$ = 128 and $\numEvi$ = 5.
The length of the concatenated sequence is thus 640 ($\LtimesM$), which is slightly longer than the maximum length of BERT's input sequence (i.e., 512 tokens).

Next, we perform sentence-level self-attention using a similar procedure.
First, we split $\G$ back into individual sequences $\Gset$.
Then, we pick the first hidden state vector from each $\G_j$, which corresponds to that of the CLS token.
Let $\F$ be the concatenation of all the first hidden state vectors $\gCLSset$.
We obtain the final representation $\E$ of the evidence sentences:
\begin{equation}
\EqResConSent.
\end{equation}
We can use $\E$ as the keys and values in Eq.~(\ref{eq:a}).
Note that we do not share the parameters among the different MHA layers.

\subsubsection*{Joint training}

\noindent
Since the sentence-selection label assigned to each evidence sentence is available at training time,
we can use it to guide our veracity prediction model.
We apply the idea of multi-task learning (MTL)~\cite{MTL93,ruder17}, in which
we consider veracity relation prediction to be our main task and evidence sentence selection to be our auxiliary task.
Our goal is to leverage training signals from our auxiliary task to improve the performance of our main task.
Note that the sentence-selection component here is independent of the stand-alone model
(i.e., our model in Section~\ref{sec:sent-selection} or an alternative model in Section~\ref{sec:results}).

\begin{figure}[!t]
  \centering
  \includegraphics[width=1.0\columnwidth]{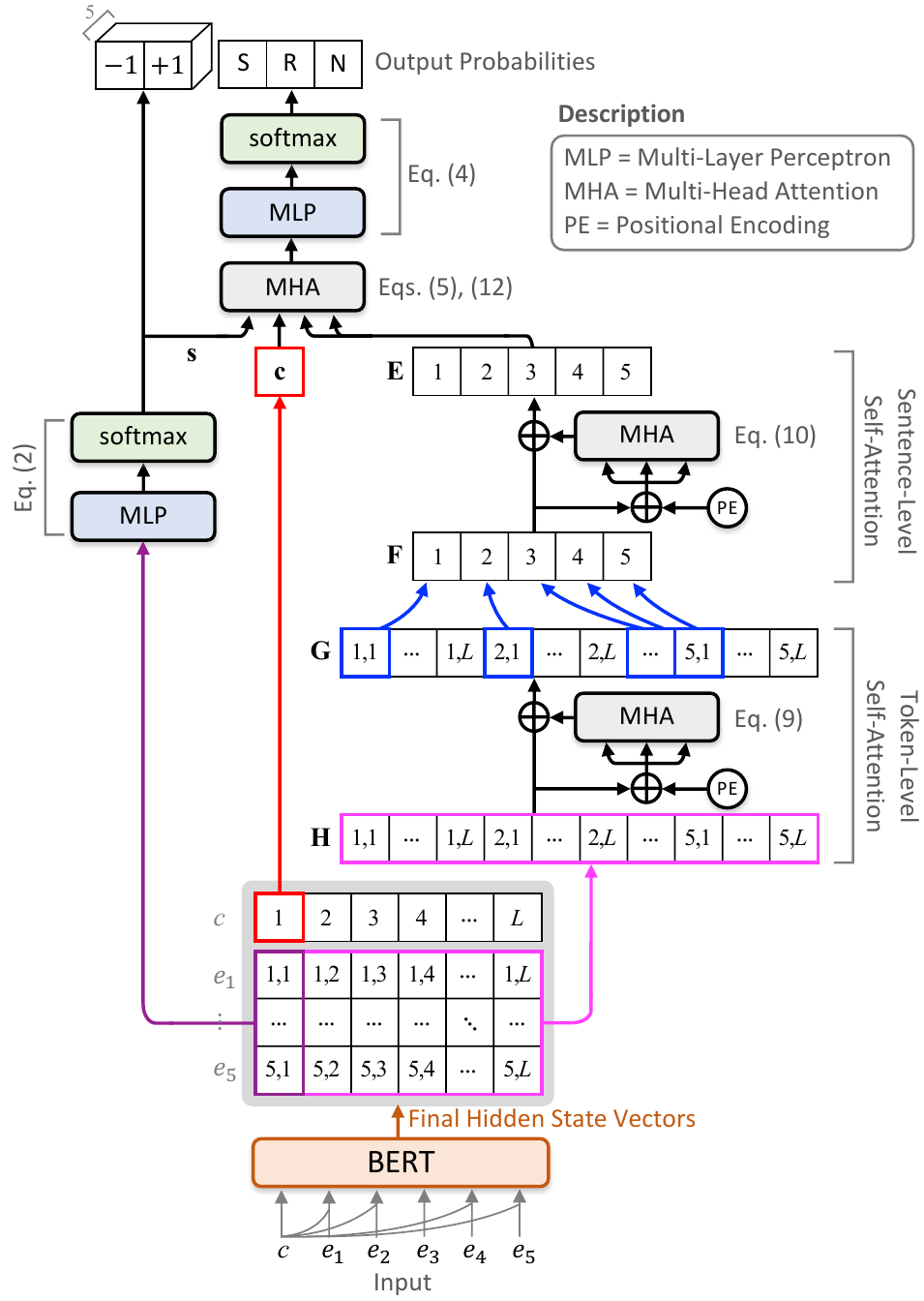}
  \caption{
    Architecture of our multi-level attention (MLA) model.
    The model takes as input a claim together with five evidence sentences.
    These sentences can be derived from any sentence selector.
    BERT encodes each sentence into a sequence of hidden state vectors, each of which is denoted by a squared box.
    The first hidden state vector (corresponding to the CLS token) is used for classification.
    MLA applies token- and sentence-level self-attentions and
    combines all the hidden state vectors as well as
    the sentence-selection scores at the final attention layer.
  }
  \label{fig:arch}
\end{figure}

Let $\scorevec$ be the vector of sentence-selection scores,
where  $s_m$ denotes the probability distribution of the positive class returned by our sentence-selection component.
We propose using $\s$ as a gate vector to determine how much of the values should be maintained before applying a residual connection followed by a linear projection.
Thus, we modify Eq.~(\ref{eq:head}) with:
{\setlength{\jot}{0.5\baselineskip}
\begin{align*}
  \EqsValueOnly
\end{align*}
}%
where $\odot$ means the element-wise multiplication.

Our modification is close to ~\newcite{Shaw18}'s method in which extra vectors are added to the keys and the values after applying the linear projections.
During development, we found that their method does not work well in our task.
We compare different strategies in Section~\ref{sec:ablation}, including applying the gate vector to the keys or both the keys and the values.

Finally, we combine Eqs.~(\ref{eq:loss-selection}) and (\ref{eq:loss-prediction}) to get our composite loss function:
\begin{equation}\label{eq:loss-joint}
  \EqLossJoint,
\end{equation}%
where $\lambda$ is the weighting factor of the sentence-selection component.

To summarize, our model, shown in Figure~\ref{fig:arch}, contains
token-level attention over a claim-evidence pair through BERT,
token- and sentence-level self-attentions across an evidence set,
and claim-evidence cross-attention incorporating the sentence-selection scores through joint training.
Hence, we call it the multi-level attention (MLA) model.

\section{Experiments}

\subsection{Dataset and evaluation metrics}

Table~\ref{tab:fever-dataset} shows the statistics of the FEVER dataset.
We used the corpus of the June 2017 Wikipedia dump, which contains 5,416,537 articles preprocessed by~\newcite{FEVER1}.
We used the document retrieval results given by~\newcite{Athene18},
containing the predicted Wikipedia article titles (i.e., document IDs)
for all the claims in the training/dev/test sets.
Following~\cite{Stammbach19,Soleimani19,KGAT20}, we prefixed the
Wikipedia article titles to the candidate sentences to alleviate the
co-reference resolution problem.

We evaluated performance by using the label accuracy (LA) and FEVER score.
LA measures the 3-way classification accuracy of veracity relation prediction.
The FEVER score reflects the performance of both evidence sentence selection and veracity relation
prediction, where a complete set of true evidence sentences is
present in the selected sentences, and the claim is correctly labeled.
We used the official FEVER scorer during development.\footnote{
\url{https://github.com/sheffieldnlp/fever-scorer}}
We limited the number of the selected sentences to five ($\numEvi$~=~5) according to the FEVER scorer.
The performance on the blind test set was evaluated through the FEVER Challenge site.

\begin{table}[t!]
\centering
{\small
\setlength\tabcolsep{3pt}
\begin{tabular}{lccc}
\toprule
Split & $\SSS$ & $\RRR$ & $\NNN$\\
\midrule
Training & 80,035 & 29,775 & 35,659\T\\
Dev      & $\enspace$6,666 & $\enspace$6,666 & $\enspace$6,666\\
Test      & $\enspace$6,666 & $\enspace$6,666 & $\enspace$6,666\B\\
\bottomrule
\end{tabular}
}
\caption{ Statistics of the FEVER dataset. Veracity relation labels and evidence sentences of
  the test set are not publicly available. }
\label{tab:fever-dataset}
\end{table}

\subsection{Training details}\label{sec:training-details}

We implemented our model on top of HuggingFace's Transformer~\cite{HFT20}.
The dimension of hidden state vectors $\dimHid$ and the number of heads $\numHead$ varied according to those of the pre-trained models.
We used BERT-base ($\dimHid$ = 786; $\numHead$ = 12) for our stand-alone sentence-selection model and tried various BERT-style models for MLA.

We trained all models using Adafactor~\cite{Adafactor18}
with a batch size of 256,
a linear learning rate decay, a warmup ratio of 0.06, and a gradient clipping of 1.0.
Following the default configuration of HuggingFace's Transformer, we
initialized all parameters by sampling from $\N(0,0.02)$ and setting
the biases to 0, except for the pre-trained models.
We set $\lambda$ in Eq.~(\ref{eq:loss-joint}) to 1.
We trained each model for 2 epochs with a learning rate of 5e-5, unless otherwise specified.

For regularization, we applied dropout~\cite{Hinton12} with a
probability of 0.1 to the MHA layers, MLP layers, and gated values in Eq.~(\ref{eq:gate}).
We computed the class weight $\beta_y$ in Eq.~(\ref{eq:loss-prediction}) by:
\begin{equation}
 \EqNormCW
\end{equation}
where $\hbeta_y$ is the balanced heuristic used in scikit-learn~\cite{sklearn11} and $\beta_y$ is normalized to sum to 1.
In our case, $N$~=~145,469 is the total number of training examples,
$|\Y|$~=~3 is the number of classes,
and $N_y$ is the number of training examples in $y$ (i.e., the first row in Table~\ref{tab:fever-dataset}).
We interpreted $\hbeta_y$ as the ratio of the balanced class distribution ($N/|\Y|$) to the observed one ($N_y$).
Here, we wanted to penalize the less-observed classes, like $\RRR$ and $\NNN$, more.

\subsection{Results}\label{sec:results}

\subsubsection*{Baselines}

The use of different pre-trained and pipeline models in the previous work makes a fair comparison difficult.
For this reason, we chose baseline models that use BERT-base for pre-training and~\newcite{Athene18}'s document retrieval results.
We designed two sets of experiments.

First, we required that all the models use the same sentence-selection model, which is~\newcite{Athene18}'s ESIM.\footnote{
We used the sentence-selection results reproduced by~\newcite{GEAR19}.}
For the veracity relation prediction,~\newcite{Athene18} incorporate ESIM with attention and pooling operations to get a representation of a claim and top five selected sentences.
\newcite{Soleimani19} make five independent predictions for each claim-evidence pair and use a heuristic~\cite{Malon18} to get a final prediction.
GEAR~\cite{GEAR19} is a graph-based model for evidence aggregating and reasoning.
KGAT~\cite{KGAT20} is a kernel graph attention model.
Second, we allowed different sentence-selection models.
\newcite{Soleimani19} use HNM to select negative examples with the highest loss values,
while our negative examples are sampled once from both the ground-truth and retrieved documents, as described in Section~\ref{sec:sent-selection}.

\begin{table}[t!]
\centering
{\small
\setlength\tabcolsep{6pt}
\begin{tabular}{lcc}
\toprule
Model & LA & FEVER\\
\midrule
\multicolumn{3}{c}{Sentence selection with ESIM}\T\\
\newcite{Athene18}                 & 68.49  & 64.74\T\\
\newcite{Soleimani19}              & 71.70  & 69.79\\
GEAR$\sdag$~\cite{GEAR19}          & 74.84  & 70.69\\
KGAT$\sdag$~\cite{KGAT20}          & 75.51  & 71.61\\
MLA~(Ours)                         & {\bf76.30}  & {\bf72.83}\B\\
\midrule
\multicolumn{3}{c}{Sentence selection with BERT-base}\T\\
\newcite{Soleimani19}$\sddag$      & 73.54  & 71.33\T\\
MLA~(Ours)                         & {\bf76.92}  & {\bf73.78}\B\\
\bottomrule
\end{tabular}
}
\caption{
LA and FEVER score results on the dev set.
All the models use the document retrieval results from \newcite{Athene18}.
Results marked with $\sdag$ indicate using ESIM with a threshold filter, and $\sddag$ indicates using BERT-base with HNM.
}
\label{tab:dev-verification-results}
\end{table}

\begin{table}[t!]
\centering
{\small
\setlength\tabcolsep{2.5pt}
\begin{tabular}{lccc}
\toprule
Model & Prec & Rec@5 & F1\\
\midrule
ESIM~\cite{Athene18}                      & 24.08 & 86.72 & 37.69\T\\
BERT-base$\sddag$~\cite{Soleimani19}      & 25.13 & 88.29 & 39.13\\
BERT-base~(Ours)                          & {\bf25.63} & {\bf88.64} & {\bf39.76}\T\\
$\enspace$ w/o sampling from retrieved docs. & 23.59 & 87.18 & 37.13\B\\
\bottomrule
\end{tabular}
}
\caption{
Sentence selection results on the dev set.
Result marked with $\sddag$ indicates using HNM.}
\label{tab:dev-selection-results}
\end{table}

Table~\ref{tab:dev-verification-results} shows the results of the two settings on the dev set.
MLA outperforms the other baselines in both settings.
Table~\ref{tab:dev-selection-results} shows the sentence-selection results returned by the FEVER scorer.
The precision, recall@5, and F1 are consistent across the three models.
\newcite{Athene18} use ESIM with a pairwise hinge loss, while~\newcite{Soleimani19} use a pointwise loss with HNM.
Our model is also a pointwise approach but simpler to train.
Without sampling non-evidence sentences from the retrieved documents, all the scores drop by around 2\%, indicating that our technique is useful.
In the following sections, we will refer to our BERT-base sentence-selection results with MLA.

\subsubsection*{Effect of pre-trained models}

The next set of experiments examined the benefits of using different pre-trained models. 
ALBERT~\cite{ALBERT20} is a lite BERT training approach that uses cross-layer
parameter sharing and replaces next sentence prediction with sentence ordering.
RoBERTa~\cite{RoBERTa19} is a robustly optimized BERT approach that introduces better training schemes, including dynamic masking, larger batch size, and other techniques.
We chose these two BERT-style models because they can be easily plugged into our implementation without much modification.

\begin{table}
\centering
{\small
\setlength\tabcolsep{6pt}
\begin{tabular}{lrcc}
\toprule
Pre-trained model & \# Params & LA & FEVER\\
\midrule
BERT-base  & 117M  & 76.92 & 73.78\T\\
BERT-large & 349M  & 77.27 & 74.10\\
ALBERT-base  & 20M & 76.58 & 73.83\T\\
ALBERT-large & 33M & 76.94 & 74.24\\
RoBERTa-base  & 132M & 77.54 & 74.41\T\\
RoBERTa-large & 370M & {\bf79.31} & {\bf75.96}\B\\
\bottomrule
\end{tabular}
}
\caption{
LA and FEVER score results of MLA on the dev set using different pre-trained models.
The second column shows the number of parameters, including those from the pre-trained model and our task-specific layers (i.e., MHA and MLP layers).
}
\label{tab:dev-pretrained-results}
\end{table}

Table~\ref{tab:dev-pretrained-results} shows the results of the different pre-trained models on the dev set.
For all the large pre-trained models, we decreased the learning rate to 2e-5 and trained them for 3 epochs. 
Additional results including training times can be found in Appendix~\ref{appendix:pretrained}.
As shown in the table, BERT and ALBERT perform similarly, while ALBERT has fewer parameters.
RoBERTa offers consistent improvements over the other two models and achieves the best performance with its large model.
Therefore, we applied MLA with RoBERTa-large to the blind test set.

\subsubsection*{Comparison with state-of-the-art methods}

\begin{table}[t!]
\centering
{\small
\setlength\tabcolsep{5pt}
\begin{tabular}{lcc}
\toprule
Model & LA & FEVER\\
\midrule
\newcite{Athene18}                   & 65.46  & 61.58\T\\
\newcite{Yoneda18}                   & 67.62  & 62.52\\
\newcite{Nie19}                      & 68.21  & 64.21\B\\
GEAR$\sdag$~\cite{GEAR19}            & 71.60  & 67.10\T\\
SR-MRS$\sdag$~\cite{SRMRS19}         & 72.56  & 67.26\\
BERT$\sddag$~\cite{Soleimani19}      & 71.86  & 69.66\\
KGAT$\sdia$~\cite{KGAT20}            & 74.07  & 70.38\\
DREAM$\sclub$~\cite{DREAM20}         & 76.85  & 70.60\\
HESM$\sspade$~\cite{HESM20}          & 74.64  & 71.48\\
CorefRoBERTa$\sdia$~\cite{Coref20}   & 75.96  & 72.30\\
MLA$\sdia$(Ours)                     & {\bf77.05}  & {\bf73.72}\B\\
\bottomrule
\end{tabular}
}
\caption{ LA and FEVER score results on the blind test set.
  Results marked with $\sdag$ indicate using BERT-base, $\sddag$
  BERT-large, $\sdia$ RoBERTa-large, $\sclub$ XLNet, and $\sspade$ ALBERT-large.}
\label{tab:test-verification-results}
\end{table}

Table~\ref{tab:test-verification-results} shows the results on the blind test set.\footnote{
The results can also be found on the FEVER leaderboard: \url{https://competitions.codalab.org/competitions/18814\#results}}
The results are divided into two groups.
The first group represents the top scores of the FEVER shared task, including those of \newcite{Athene18,Yoneda18,Nie19}.
The second group contains recently published results after the shared task.
GEAR~\cite{GEAR19}, KGAT~\cite{KGAT20}, and DREAM~\cite{DREAM20} are graph-based models.
SR-MRS~\cite{SRMRS19} uses a semantic retrieval module for selecting evidence sentences.
HESM~\cite{HESM20} uses a multi-hop evidence retriever and a hierarchical evidence aggregation model.
\mbox{CorefRoBERTa}~\cite{Coref20} trains KGAT by using a pre-trained model that combines a co-reference prediction loss.
Their pre-trained model is initialized with RoBERTa-large's parameters
and further trained on Wikipedia.
MLA outperforms all the published models and yields 1.09\% and 1.42\% improvements in LA and FEVER score, respectively, over \mbox{CorefRoBERTa}.
Additional sentence-selection results can be found in Appendix~\ref{appendix:sent}.

\subsection{Ablation study}\label{sec:ablation}

\begin{table}[t]
\centering
{\small
\setlength\tabcolsep{6pt}
\begin{tabular}{lcc}
\toprule
Model & LA & FEVER\\
\midrule
MLA (full)                                  & {\bf76.92}  & {\bf73.78}\T\\
\enspace w/o token-level self-attention     & 76.30 & 73.20 \\
\enspace w/o sentence-level self-attention  & 76.50 & 73.41 \\
\enspace w/o class weighting                & 76.44 & 73.14 \\
\enspace w/o joint training                 & 76.65 & 73.22\B\\
\bottomrule
\end{tabular}
}
\caption{Ablation studies of the proposed components on the dev set with BERT-base.}
\label{tab:ablation-comp}
\end{table}

\begin{table}[t]
\centering
{\small
\setlength\tabcolsep{6pt}
\begin{tabular}{lcc}
\toprule
Model & LA & FEVER\\
\midrule
MLA (w/ value)                 & {\bf76.92}  & {\bf73.78}\T\\
\enspace w/ key                & 76.74 & 73.65\\
\enspace w/ key \& value       & 76.82 & 73.60\\
\enspace w/ dot-product        & 76.70 & 73.51\\
\enspace w/o using $\s$        & 76.64 & 73.47\B\\
\bottomrule
\end{tabular}
}
\caption{Ablation studies of different strategies for using the sentence-selection scores $\s$ on the dev set with BERT-base.}
\label{tab:ablation-s}
\end{table}

We conducted two sets of ablation studies on the dev set using MLA with BERT-base.
First, we examined the effect of our proposed components.
Table~\ref{tab:ablation-comp} shows that all the components contribute to the final results.
Without class weighting, Eq.~(\ref{eq:loss-prediction}) falls back to the standard cross-entropy loss.
Without joint training, MLA is a stand-alone veracity prediction model.
These results suggest that token-level self-attention and class weighting are the two most important components of our model.

Second, we explored a number of strategies for exploiting the sentence-selection scores $\s$.
MLA basically uses $\s$ as a gate vector and only applies it to the values, as described in Eq.~(\ref{eq:gate}).
We can apply the same calculation to the keys or both the keys and the values.
In addition, we can use $\s$ as a bias vector and add it to the scaled dot-product term, as done by~\newcite{Yang18}.
Table~\ref{tab:ablation-s} shows the results of the aforementioned strategies.
These results indicate that applying $\s$ to the values produces the best results.

\subsection{Error analysis}

\begin{table}[t]
\centering
{\small
\begin{tabular}{p{0.14\columnwidth}p{0.76\columnwidth}}
\toprule
{\bf ID:}        & 35237\T\\
{\bf Claim:}      & Philomena is a film nominated for seven awards.\\%
{\bf Evidence:}  & [\textit{Philomena\_(film)}] It was also nominated for four BAFTA Awards and three Golden Globe Awards.$^{9}$\\
\multicolumn{2}{l}{{\bf Annotated label:} $\SSSC$} \\
\multicolumn{2}{l}{{\bf Predicted label:} $\RRRC$} \\
\multicolumn{2}{c}{(a)}\B\\
\midrule
{\bf ID:}       & 33547\T\\
{\bf Claim:}    & Mick Thomson was born in Ohio.\\
{\bf Evidence:} & [\textit{Mick\_Thomson}] Born in Des Moines, Iowa, he is best known as one of two guitarists in Slipknot, in which he is designated \#7.$^1$\\
\multicolumn{2}{l}{{\bf Annotated label:} $\SSSC$} \\
\multicolumn{2}{l}{{\bf Predicted label:} $\RRRC$}\\
\multicolumn{2}{c}{(b)}\B\\
\midrule
{\bf ID:}       & 73443 \\
{\bf Claim:}    & Heavy Metal music was developed in the United Kingdom.\\
{\bf Evidence:} & [\textit{Heavy\_metal\_music}] Heavy metal (or simply metal) is a genre of rock music that developed in the late 1960s and early 1970s, largely in the United Kingdom and the United States.$^0$\\
\multicolumn{2}{l}{{\bf Annotated label:} $\RRRC$} \\
\multicolumn{2}{l}{{\bf Predicted label:} $\SSSC$} \\
\multicolumn{2}{c}{(c)}\B\\
\midrule
{\bf ID:}       & 212780\T\\
{\bf Claim:}    & Harvard University is the first University in the U.S. \\
{\bf Evidence:} & [\textit{Harvard\_University}] Established originally by the Massachusetts legislature and soon thereafter named for John Harvard (its first benefactor), Harvard is the United States' oldest institution of higher learning ...$^3$\\
\multicolumn{2}{l}{{\bf Annotated label:} $\SSSC$} \\
\multicolumn{2}{l}{{\bf Predicted label:} $\NNNC$} \\
\multicolumn{2}{c}{(d)}\B\\
\bottomrule
\end{tabular}
}
\caption{
Examples where the models disagree with the annotated labels.
}
\label{tab:errors}
\end{table}

To better understand the limitations of our method, we manually
inspected 100 prediction errors on the dev set, where the true
evidence sentences are present in the predicted sentences but MLA
failed to predict the veracity relation labels.
Here, we required that both BERT-base and RoBERTa-large MLA models produce the same errors.

Table~\ref{tab:errors}(a) shows a prediction error requiring complex reasoning that our models are unable to deal with.
The claim ``\textit{Philomena is a film nominated for \textbf{seven} awards.}'' is supported
by the evidence ``\textit{It was also nominated for \textbf{four} BAFTA Awards and \textbf{three} Golden Globe Awards.}''.
In this case, the models must understand that four plus three equals seven.

Table~\ref{tab:errors}(b) shows a possible annotation error.
The claim ``\textit{Mick Thomson was born in \textbf{Ohio}.}'' is annotated as $\SSSC$,
while the evidence ``\textit{Born in Des Moines, \textbf{Iowa}, he is best known as ...}'' refutes the claim.
Our models also predict $\RRRC$.

Table~\ref{tab:errors}(c) shows the half-true claim
``\textit{Heavy Metal music was developed in \textbf{the United Kingdom}.}'', which is annotated as $\RRRC$.
However, the evidence ``\textit{Heavy metal (or simply metal) is ... developed ... in \textbf{the United Kingdom and the United States}.}'' would indicate that the claim is partly true.
The half-true label is defined in some previous smaller datasets~\cite{vlachos14,wang17}, but not in the FEVER dataset.

Table~\ref{tab:errors}(d) shows the questionable claim
``\textit{Harvard University is \textbf{the first University} in the U.S.}'',
which is annotated as $\SSSC$ by the evidence
``\textit{... Harvard is the United States’ \textbf{oldest institution} of higher learning ...}''.
However, this evidence does not directly support the claim.\footnote{The topic is still under debate:
\url{https://en.wikipedia.org/wiki/First\_university\_in\_the\_United\_States}.}
Our models predict $\NNNC$.
Our analysis results suggest that probing disagreements between an ensemble of models and annotators may help improve annotation consistency.
Additional results on error analysis are given in Appendix~\ref{appendix:error}.

\section{Conclusion}

We have presented a multi-level attention model that operates on linear sequences.
We find that, when trained properly, the model outperforms its graph-based counterparts.
Our results suggest that a sequence model is sufficient and can serve as a strong baseline.
Using better upstream components (i.e., a better document retriever or sentence selector) or larger pre-trained models would further improve the performance of our model.
Training models that are robust to adversarial examples while maintaining high performance for normal ones is an important direction for our future work.

\section*{Acknowledgments}

We thank Erica Cooper (NII) for providing valuable feedback on an earlier draft of this paper.
This work is supported by JST CREST Grants (JPMJCR18A6 and JPMJCR20D3) and MEXT KAKENHI Grants (21H04906), Japan.

\bibliographystyle{acl_natbib}
\bibliography{acl2021}

\appendix

\section{Additional results on different pre-trained models}\label{appendix:pretrained}

Table~\ref{tab:dev-add-pretrained-results} shows the results of different pre-trained models in detail.
All the pre-trained models used in our experiments also come from HuggingFace.\footnote{
\url{https://huggingface.co/transformers/pretrained_models.html}.}
We conducted each experiment on a single NVIDIA Tesla A100 GPU with 40 GB RAM.
We used a batch size of 256 with gradient accumulation to control memory.

\begin{table*}
\centering
{\small
\begin{tabular}{lrccrcc}
\toprule
Pre-trained model & \# Params & Learning rate & Epochs & Time & LA & FEVER\\
\midrule
BERT-base  & 117M & 5e-5 & 2 & $\phantom{0h}$ 46m  & 76.92 & 73.78\T\\
BERT-large & 349M & 2e-5 & 3 & 2h  50m             & 77.27 & 74.10\\
ALBERT-base  & 20M & 5e-5 & 2 & $\phantom{0h}$ 57m & 76.58 & 73.83\T\\
ALBERT-large & 33M & 2e-5 & 3 & 3h  35m & 76.94 & 74.24\\
RoBERTa-base  & 132M & 5e-5 & 2 & $\phantom{0h}$ 45m & 77.54 & 74.41\T\\
RoBERTa-large & 370M & 2e-5 & 3 & 2h 49m & {\bf79.31} & {\bf75.96}\B\\
\bottomrule
\end{tabular}
}
\caption{
Additional results of MLA on the dev set using different pre-trained models.
}
\label{tab:dev-add-pretrained-results}
\end{table*}

\begin{table*}
\centering
{\small
\begin{tabular}{lllccc}
\toprule
Model  & Loss & Pre-trained model & Prec & Rec@5 & F1\\
\midrule
\newcite{Athene18}           & Pairwise           & $\qquad$--       & --    & --    & 36.97\T\\
\newcite{Yoneda18}           & Pointwise          & $\qquad$--       & --    & --    & 34.97\\
\newcite{Nie19}              & Pointwise          & $\qquad$--       & --    & --    & 52.96\B\\
\newcite{GEAR19}                 & Pairwise \& filtering     &  $\qquad$--      & --    & --    & 36.87\T\\
\newcite{SRMRS19}                & Pointwise                & BERT-base        & --    & --    & 74.62\\
\newcite{Soleimani19}            & Pointwise \& HNM          & BERT-base        & --    & --    & 38.61\\
\newcite{KGAT20}                 & Pairwise                 & BERT-base        & 25.21 & 87.47 & 39.14\\
\newcite{DREAM20}                & Pointwise                & RoBERTa          & 25.63 & 85.57 & 39.45\\
\newcite{HESM20}                 & Pointwise \& multi-hop    & ALBERT-base      & --    & --    & 52.78\\
\newcite{Coref20}                & (adopting \newcite{KGAT20}'s results)  & $\qquad$--       & --    & --    &  39.14\\
This work                             & Pointwise                & BERT-base        & 25.33 & 87.58 & 39.29\B\\
\bottomrule
\end{tabular}
}
\caption{
  Sentence-selection results on the blind test set.
  The F1 results can be found on the FEVER leaderboard: \url{https://competitions.codalab.org/competitions/18814\#results}.
}
\label{tab:test-selection-results}
\end{table*}

\section{Additional sentence-selection results}\label{appendix:sent}

Table~\ref{tab:test-selection-results} shows the results of various sentence-selection models on the test set.
Not all published models report precision and recall.
Our precision, recall@5, and F1 scores are slightly better than those of \newcite{KGAT20}.
Our sentence-selection model took 1 hour and 10 minutes to train.
We find that getting high recall in evidence sentence selection is necessary to achieve good results in veracity relation prediction.

\section{Additional error analysis}\label{appendix:error}

Here, we provide additional examples of errors, including
complex reasoning errors (Table~\ref{tab:ana-complex}),
possible annotation errors (Table~\ref{tab:ana-anno}),
half-true claims (Table~\ref{tab:ana-half}), and
questionable claims (Table~\ref{tab:ana-questionable}).

\begin{table}[h]
\centering
{\small
\begin{tabular}{p{0.14\columnwidth}p{0.76\columnwidth}}
\toprule
{\bf ID:}       & 112396\T\\
{\bf Claim:}    & Aristotle spent the majority of his life in Athens.\\%
{\bf Evidence:}  & [\textit{Aristotle}] At seventeen or eighteen years of age, he joined Plato's Academy in Athens and remained there until the age of thirty-seven (c. 347 BC).$^{2}$\\
\multicolumn{2}{l}{{\bf Annotated label:} $\SSSC$} \\
\multicolumn{2}{l}{{\bf Predicted label:} $\RRRC$}\B\\
\midrule
{\bf ID:}       & 3111\T\\
{\bf Claim:}    & Luis Fonsi was born in the eighties.\\%
{\bf Evidence:}  & [\textit{Luis\_Fonsi}] Luis Alfonso Rodr\'iguez L\'opez-Cepero, more commonly known by his stage name Luis Fonsi, (born April 15, 1978) is a Puerto Rican singer, songwriter and actor.$^{0}$\\
\multicolumn{2}{l}{{\bf Annotated label:} $\RRRC$} \\
\multicolumn{2}{l}{{\bf Predicted label:} $\SSSC$}\B\\
\midrule
{\bf ID:}       & 64685\T\\
{\bf Claim:}    & The Bassoon King is the full title a book.\\%
{\bf Evidence:}  & [\textit{The\_Bassoon\_King}] The Bassoon King: My Life in Art, Faith, and Idiocy is a non-fiction book authored by American actor Rainn Wilson.$^{0}$\\
\multicolumn{2}{l}{{\bf Annotated label:} $\RRRC$} \\
\multicolumn{2}{l}{{\bf Predicted label:} $\SSSC$}\B\\
\midrule
{\bf ID:}       & 102001\T\\
{\bf Claim:}    & Jens Stoltenberg was Prime Minister of Norway once.\\%
{\bf Evidence:}  & [\textit{Jens\_Stoltenberg}] Stoltenberg served as Prime Minister of Norway from 2000 to 2001 and from 2005 to 2013.$^{4}$\\
\multicolumn{2}{l}{{\bf Annotated label:} $\RRRC$} \\
\multicolumn{2}{l}{{\bf Predicted label:} $\SSSC$}\B\\
\bottomrule
\end{tabular}
}
\caption{
Examples of prediction errors requiring complex reasoning.
}
\label{tab:ana-complex}
\end{table}

\begin{table}[h]
\centering
{\small
\begin{tabular}{p{0.14\columnwidth}p{0.76\columnwidth}}
\toprule

{\bf ID:}        &  117520\T\\
{\bf Claim:}    &  The host of The Joy of Painting was Bob Ross.\\
{\bf Evidence:} & [\textit{Bob\_Ross}] He was the creator and host of The Joy of Painting, an instructional television program that aired from 1983 to 1994 ...$^1$ \\
\multicolumn{2}{l}{{\bf Annotated label:} $\RRRC$} \\
\multicolumn{2}{l}{{\bf Predicted label:} $\SSSC$} \\
\midrule
{\bf ID:}       &  114640\T\\
{\bf Claim:}    &  IMDb is not user-edited.\\
{\bf Evidence:} &  [\textit{IMDb}] The site enables registered users to submit new material and edits to existing entries.$^{10}$ \\
\multicolumn{2}{l}{{\bf Annotated label:} $\SSSC$} \\
\multicolumn{2}{l}{{\bf Predicted label:} $\RRRC$}\B\\
\midrule
{\bf ID:}        & 137678\T\\
{\bf Claim:}    & Food Network is available to approximately 96,931,000 pay television citizens. \\
{\bf Evidence:} & [\textit{Food\_Network}] As of February 2015, Food Network is available to approximately 96,931,000 pay television households ...$^8$ \\
\multicolumn{2}{l}{{\bf Annotated label:} $\RRRC$} \\
\multicolumn{2}{l}{{\bf Predicted label:} $\SSSC$}\B\\
\midrule
{\bf ID:}       & 34195\T\\
{\bf Claim:}    & Annie Lennox was named ``The Greatest White Soul Singer Alive'' by VH1. \\
{\bf Evidence:} & [\textit{Annie\_Lennox}] Lennox has been named ``The Greatest White Soul Singer Alive'' by VH1 ...$^{19}$\\
\multicolumn{2}{l}{{\bf Annotated label:} $\RRRC$} \\
\multicolumn{2}{l}{{\bf Predicted label:} $\SSSC$}\B\\
\bottomrule
\end{tabular}
}
\caption{
Example of possible annotation errors.
}
\label{tab:ana-anno}
\end{table}

\begin{table}[h]
\centering
{\small
\begin{tabular}{p{0.14\columnwidth}p{0.76\columnwidth}}
\toprule
{\bf ID:}       & 174029\T\\
{\bf Claim:}    & The Endless River came out in 1995 and is Pink Floyd's fifteenth studio album.\\
{\bf Evidence:} & [\textit{The\_Endless\_River}] The Endless River is the fifteenth and final studio album by the English rock band Pink Floyd.$^0$\\
\multicolumn{2}{l}{{\bf Annotated label:} $\RRRC$} \\
\multicolumn{2}{l}{{\bf Predicted label:} $\SSSC$}\B\\
\midrule
{\bf ID:}       & 161094 \\
{\bf Claim:}    & French Indochina was a grouping of territories.\\
{\bf Evidence:} & [\textit{French\_Indochina}] French Indochina (previously spelled as French Indo-China) ...  was a grouping of French colonial territories in Southeast Asia.$^0$ \\
\multicolumn{2}{l}{{\bf Annotated label:} $\RRRC$} \\
\multicolumn{2}{l}{{\bf Predicted label:} $\SSSC$} \\
\midrule
{\bf ID:}       & 48148\T\\
{\bf Claim:}    & On Monday August 19, 1945, Ian Gillan was born.\\
{\bf Evidence:} & [\textit{Ian\_Gillan}] Ian Gillan (born 19 August 1945) is an English singer and songwriter.$^0$\\
\multicolumn{2}{l}{{\bf Annotated label:} $\SSSC$} \\
\multicolumn{2}{l}{{\bf Predicted label:} $\NNNC$} \\
{\bf Note:} & August 19, 1945 is Sunday, not Monday.\B\\
\midrule
{\bf ID:}       & 85350\T\\
{\bf Claim:}    & Andrew Kevin Walker was born on Monday August 14, 1964.\\
{\bf Evidence:} & [\textit{Andrew\_Kevin\_Walker}] Andrew Kevin Walker (born August 14, 1964) is an American BAFTA-nominated screenwriter .$^0$\\
\multicolumn{2}{l}{{\bf Annotated label:} $\SSSC$} \\
\multicolumn{2}{l}{{\bf Predicted label:} $\NNNC$} \\
{\bf Note:} & August 19, 1945 is Friday, not Monday.\B\\
\bottomrule
\end{tabular}
}
\caption{
Examples of half-true claims.
}
\label{tab:ana-half}
\end{table}

\begin{table}[h]
\centering
{\small
\begin{tabular}{p{0.14\columnwidth}p{0.76\columnwidth}}
\toprule
{\bf ID:}       & 92900\\
{\bf Claim:}    & The Indian Institute of Management Bangalore offers a business executive training program.\\
{\bf Evidence:} & [\textit{Indian\_Institute\_of\_Management\_Bangalore}] It offers Post Graduate, Doctoral and executive training programmes.$^5$\\
\multicolumn{2}{l}{{\bf Annotated label:} $\SSSC$} \\
\multicolumn{2}{l}{{\bf Predicted label:} $\NNNC$} \\
{\bf Note:} & The evidence does not specify that the institute offers a \textit{business} executive training program.\B\\
\midrule
{\bf ID:}       & 46271\T\\
{\bf Claim:}    & Prescott, Arizona is in northern Yavapai County.\\
{\bf Evidence:} & [\textit{Prescott,\_Arizona}] Prescott ... is a city in Yavapai County, Arizona, United States.$^0$\\
\multicolumn{2}{l}{{\bf Annotated label:} $\SSSC$} \\
\multicolumn{2}{l}{{\bf Predicted label:} $\NNNC$} \\
{\bf Note:} & The evidence does not specify that Prescott is in the \textit{northern} part of Yavapai County.\B\\
\midrule
{\bf ID:}       & 227779\T\\
{\bf Claim:}    & Lyon is a city in Southwest France.\\
{\bf Evidence:} & [\textit{Lyon}] Lyon had a population of 506,615 in 2014 and is France's third-largest city after Paris and Marseille.$^4$\\
\multicolumn{2}{l}{{\bf Annotated label:} $\SSSC$} \\
\multicolumn{2}{l}{{\bf Predicted label:} $\RRRC$} \\
{\bf Note:} & The evidence does not directly support the claim.\B\\
\bottomrule
\end{tabular}
}
\caption{
Examples of questionable claims.
}
\label{tab:ana-questionable}
\end{table}

\end{document}